\pdfoutput=1
\documentclass[conference]{IEEEtran}
\IEEEoverridecommandlockouts
\usepackage{cite}
\usepackage{amsmath,amssymb,amsfonts}
\usepackage{algorithmic}
\usepackage{graphicx}
\usepackage{textcomp}
\usepackage{xcolor}
\usepackage{mathrsfs} 
\usepackage{tabu}
\usepackage{booktabs}
\usepackage{subfigure}
\usepackage{multirow}
\usepackage{makecell}
\usepackage{stmaryrd}
\usepackage{bm}
\usepackage{bbding}
\def\BibTeX{{\rm B\kern-.05em{\sc i\kern-.025em b}\kern-.08em
    T\kern-.1667em\lower.7ex\hbox{E}\kern-.125emX}}
\begin{document}
\bibliographystyle{IEEEtran}

\title{Multi-level Contrast Network for Wearables-based Joint Activity Segmentation and Recognition}

\author{\IEEEauthorblockN{Songpengcheng Xia\IEEEauthorrefmark{1}, Lei Chu\IEEEauthorrefmark{1}, Ling Pei\IEEEauthorrefmark{1}, Wenxian Yu\IEEEauthorrefmark{1}, Robert C. Qiu\IEEEauthorrefmark{2}
}
\IEEEauthorblockA{\IEEEauthorrefmark{1}School of Electronic Information and Electrical Engineering, Shanghai Jiao Tong University, Shanghai, China}
\IEEEauthorblockA{\IEEEauthorrefmark{2}School of Electronic Information and Communications, Huazhong University of Science and Technology, Wuhan, China}

}

\maketitle

\begin{abstract}
Human activity recognition (HAR) with wearables is promising research that can be widely adopted in many smart healthcare applications. In recent years, the deep learning-based HAR models have achieved impressive recognition performance. However, most HAR algorithms are susceptible to the multi-class windows problem that is essential yet rarely exploited. In this paper, we propose to relieve this challenging problem by introducing the segmentation technology into HAR, yielding joint activity segmentation and recognition. Especially, we introduce the Multi-Stage Temporal Convolutional Network (MS-TCN) architecture for sample-level activity prediction to joint segment and recognize the activity sequence. Furthermore, to enhance the robustness of HAR against the inter-class similarity and intra-class heterogeneity, a multi-level contrastive loss, containing the sample-level and segment-level contrast, has been proposed to learn a well-structured embedding space for better activity segmentation and recognition performance. Finally, with comprehensive experiments, we verify the effectiveness of the proposed method on two public HAR datasets,  achieving significant improvements in the various evaluation metrics. 
\end{abstract}


\section{Introduction}
With the development of the Internet of Things (IoT) and E-health, human activity recognition is fundamental research for various smart health applications \cite{chen2020creativebioman}. In terms of sensing modalities, HAR can be mainly divided into three categories: vision-based \cite{ gammulle2021tmmf}, radio-based \cite{ahuja2021vid2doppler} and wearables-based \cite{pei2020mars, zhang2021open}. Considering the users' privacy and the restrictions on the scope of activities, the activity recognition schemes using wearables are widely used in various applications, such as mental health pre-diagnosis \cite{chu2021ahed} and health status monitoring\cite{chen2017wearable}. 

Artificial intelligence and deep learning technology have boosted the HAR in many aspects and achieved impressive performance in recent years \cite{ma2019attnsense, chen2021deep, xia2021learning}. The researchers  utilized deep neural networks(DNNs) \cite{radu2018multimodal}, convolutional neural networks(CNNs) \cite{yang2015deep}, recurrent neural networks(RNNs) \cite{guan2017ensembles} or hybrid structures \cite{ordonez2016deep} to extract features from the related time series and then  effectively recognized corresponding activities. However, two fundamental and challenging issues hinder applying HAR algorithms in practice. 

The first one is the multi-class windows problem 
\cite{yao2018efficient, ma2019attnsense, xia2022boundary}. The sliding window scheme \cite{yao2018efficient, ma2019attnsense, chen2021deep} is a popular data preprocessing strategy widely used in HAR algorithms. With a fixed-size sliding window, the activities data will be decomposed into time series slices with a unique label in each slice. However, due to the complexity of activities in practice, such a scheme may treat multi-activities into one window, leading to wrong labeling for the activities. 
A natural way to address this tricky multi-class windows problem is to borrow experience from existing methods. For example, current methods utilized some famous image segmentation (e.g., Fully Convolutional Network (FCN) \cite{long2015fully} and U-net \cite{ronneberger2015u}) to predict every sample's label in the wearable sequence with the sample-wise cross-entropy loss, instead of predicting the activity label of the sliding window. 

Another challenge is inter-class similarity and intra-class heterogeneity in activities data, where the latent embedding space in a neural network may not be well structured.
Recently, the contrastive learning has been widely used to address this issue \cite{chen2020simple, oord2018representation}. The goal of contrastive learning is to learn a structured embedding space by comparing the anchor with positive and negative samples \cite{khaertdinov2021deep, wang2021exploring}. Oord et al. \cite{oord2018representation} proposed an auto-regressive model, called contrastive predictive coding, which could learn effective representations of different data modalities. Moreover, \cite{khosla2020supervised} extended the self-supervised contrastive method to the fully-supervised setting. Wang et al. \cite{wang2021exploring} applied the supervised contrastive loss to the image segmentation task, which makes the pixel embeddings with the same semantic class more compact. 
In \cite{deldari2021time}, Deldari et al. proposed a  self-supervised contrastive predictive coding method to detect the change point, which employed contrastive learning on the embeddings of adjacent and separated time-series segments. Moreover, \cite{eldele2021time} was proposed to learn the time-series representation from unlabeled data by using a contextual contrast module with weak and strong augmentations.

Therefore, inspired by the novel designs in video action recognition \cite{farha2019ms, gammulle2021tmmf}, semantic image segmentation \cite{wang2021exploring} and contrastive learning, our work will apply the multi-stage temporal convolutional network (MS-TCN) and contrastive learning to the wearables-based human activity segmentation and recognition. First, to avoid the multi-class windows problem in wearables-based HAR, we introduce the MS-TCN architecture to predict the sample-level activity labels, which could extract more effective representations containing enough context and obtain robust recognition performance. Furthermore, we design a multi-level contrastive loss to learn a more structured embedding space for coping with the inter-class similarity and intra-class heterogeneity.

Our contributions are listed as follows:
\begin{enumerate}
	\item We propose a novel sample-level activity prediction framework that integrates MS-TCN architecture with sample-wise cross-entropy and contrastive loss for alleviating the multi-class windows problem. 
	\item  To obtain a well-structured embedding space, we propose a multi-level contrastive loss with a hybrid hard example sampling strategy containing the sample-level and segment-level contrast, relieving the pain of inter-class similarity and intra-class heterogeneity.
	\item On public HAR datasets, we compare our proposed approach with various state-of-the-art methods. The extensive quantitative and qualitative experiments and ablation study demonstrate the promising performance of our proposed method.
\end{enumerate}

\section{Preliminaries}
\label{pre}
\subsection{Problem Definition}
In this paper, our goal is to predict the sample-level activity labels in a wearable sensory sequence. For this problem, let $S = [s_1,...,s_T], s_t \in \mathcal{R}^D$ be a $D$ dimensional input data and $Y = [y_1,...,y_T], y_t \in \mathcal{R}^C$ be the corresponding activity labels, where $T$ is the total number of sensory samples. In the inference stage, our model would directly map the input sequence $S$ to the sample-level activity labels $\hat{Y}$, instead of predicting the sliding windows' activity labels.
\begin{figure*}[h]
	\centering
	\includegraphics[width=16cm]{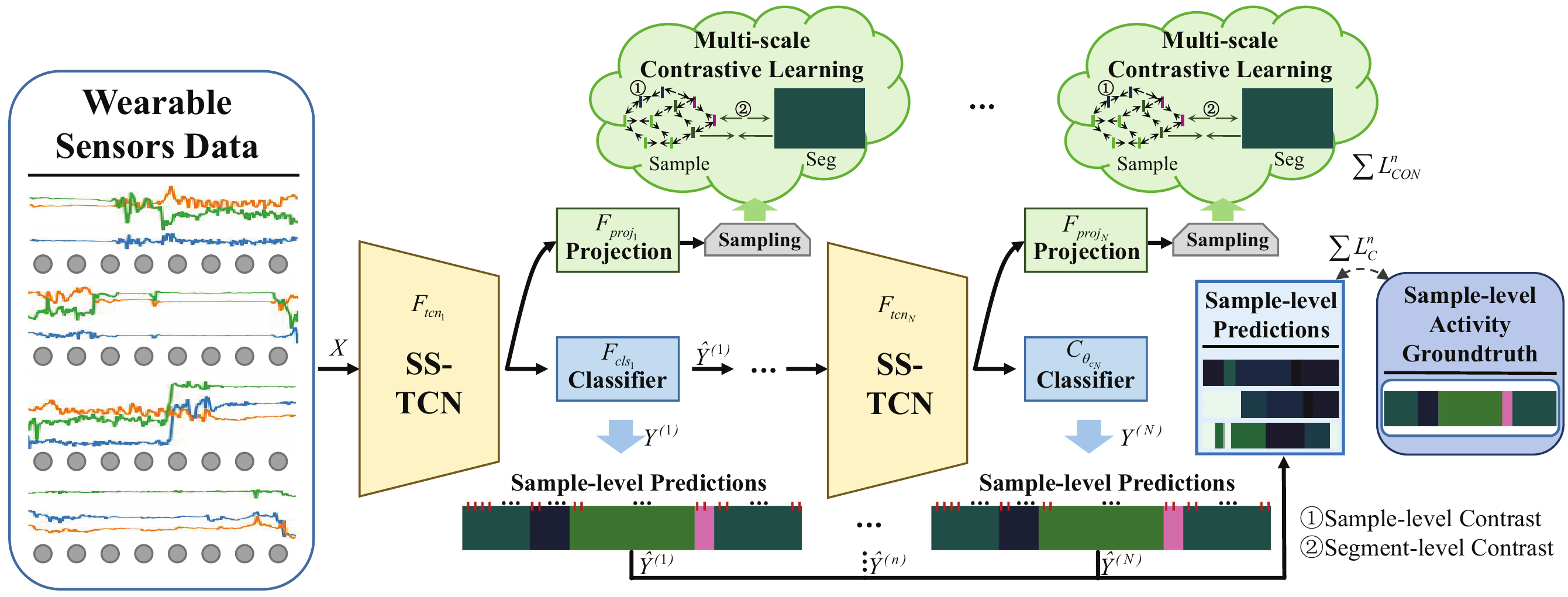}
	\caption{Illustration of our proposed multi-level contrast network for wearables-based joint activity segmentation and recognition.}
	\label{Overview}
\vspace{-0.5cm}
\end{figure*}

\subsection{Multi-Stage Temporal Convolutional Network}
MS-TCN (Multi-Stage Temporal Convolutional Network), proposed by Farha et al. \cite{farha2019ms}, is formed by stacking multiple SS-TCNs (Single-Stage Temporal Convolutional Network), which consists of dilated convolutional layers and residual layers with a dilation factor ($2^i$ for the $i^{th}$ layer). For the output ($H_{l}$) of the SS-TCN's $i^{th}$ layer, the operation with the previous layer's output ($H_{l-1}$) can be described by:
\begin{align}
\label{tcn}
\hat{H}_l = ReLU(W_d * H_{l-1} + b_d) \\
H_l = H_{l-1} + W_r * \hat{H}_l + b_r,
\vspace{-0.5cm}
\end{align}%
where $W_d$, $W_r$ are the weights of the dilated convolutional filters and a $1\times 1$ convolution, $b_d, b_r$ are the bias vectors and $*$ is the convolution operator. 

\subsection{Contrastive Learning with InfoNCE}
Contrastive learning is widely used in self-supervised learning, which aims to learn a robust representation by contrasting the anchor ($S$) with positives ($\mathcal{P}_{S}$) and negatives ($\mathcal{N}_{S}$). In a self-supervised manner, the positive sample is usually set as the anchor augmented results, and the negative samples are randomly sampled in the mini-batch. With the theory of mutual information, InfoNCE loss \cite{van2018representation} was proposed in Contrastive Predictive Coding (CPC) and widely used in various self-supervised learning tasks, which can be represented as follows:
\begin{align}
\label{info}
\resizebox{.85\linewidth}{!}{$
L^{NCE} = -\log \frac{\exp{(\boldsymbol{v \cdot v^{+}} / \tau)}}{\exp{(\boldsymbol{v \cdot v^{+}} / \tau)} + \sum\limits_{v^{-} \in \mathcal{N}_S} \exp{(\boldsymbol{v \cdot v^{-}} / \tau)}}
$},
\vspace{-0.5cm}
\end{align}%
where $\boldsymbol{v}, \boldsymbol{v^+}, \boldsymbol{v^-}$ are the feature embedding of the anchor, the positives and negatives. $\mathcal{N}_S$ is a set of negatives' embedding, and $\tau$ is the temperature hyper-parameter. 

\section{Method}
In this section, we will introduce our designed method for wearable-based human activity segmentation and recognition in detail. Fig. \ref{Overview} illustrates the overview of our approach.

\subsection{Long-term Temporal Feature Extraction with Sample-wise Cross-entropy Loss}
To address the multi-class windows problem in the traditional wearable-based HAR methods with fixed-size sliding windows, we design a novel deep learning model to predict the sample-level activity labels, instead of the windows' labels, inspired by \cite{yao2018efficient}. However, the single sample in the wearable sequence could not contain enough information for activity recognition, and the FCN adopted in \cite{yao2018efficient} might be challenging to model effective context among the samples in a wearable sequence for its spatial invariance characteristic \cite{wang2021exploring}. Therefore, we employ the single-stage temporal convolutional network as a feature extraction module to learn our encoded representation for sample-level activity prediction, introduced in Section \ref{pre}-B.

For the joint activity segmentation and recognition task, our model uses the SS-TCN module ($F_{tcn}(\cdot)$) to make each sample ($s_i$) map into encoded representation ($z_i \in Z$) and classify it into an activity category ($\hat{y}_i$) by a classifier ($F_{cls}(\cdot)$). With the sample-level ground-truth labels ($y_i$), the sample-wise classification loss can be represented by:
\begin{equation}
\begin{aligned}
\label{cls}
\resizebox{.9\linewidth}{!}{$
L_C = \sum\limits_{t=1}^{T} l(F_{cls}(F_{tcn}(s_t)), y_t)= \sum\limits_{t=1}^{T} l(F_{cls}(z_t), y_t) = \sum\limits_{t=1}^{T} l(\hat{y}_t, y_t)
$}, 
\end{aligned}%
\end{equation}
where $l(\cdot)$ represents the sample-wise cross-entropy loss. 

Following \cite{yao2018efficient, xia2022boundary}, wearable-based human activity segmentation and recognition task would be treated as a dense classification problem, whose outputs are the \textit{sample-level activity predictions}. The objective function can be optimized by minimizing $L_C$ to get the sample-level activity predictions. However, there are some disadvantages when only using the single-stage structure with sample-wise cross-entropy loss for our task: 1) only adopting the single-stage structure may lead to the fine-grained information lost; 2) this objective function with sample-wise cross-entropy loss is not sufficient to learn a structured embedding space for more robust sample-level activity recognition. Therefore, we introduce the multi-level supervised contrastive loss and multi-stage sample-level activity prediction architecture to solve these challenges.

\subsection{Learning Structured Representation with Multi-level Contrast-based Regularization}
The sample-wise cross-entropy loss may cause the fluctuation of sample-level activity predictions and make the learned embedding representation unstructured. Inspired by the supervised contrastive loss \cite{wang2021exploring, khosla2020supervised}, we design a multi-level contrastive loss to learn a more structured representation for advancing segmentation and recognition performance. 

Based on the supervised contrastive loss Eq.(\ref{info}) \cite{wang2021exploring, khosla2020supervised}, we desire to employ it for the wearable samples to form a sample-level contrastive loss, which can pull together the samples with the same class and push apart the heterogeneous samples. For the samples ($S = [s_1,...,s_T]$) in wearable sequence, with the feature extractor $F_{tcn}(\cdot)$ and the projection head $F_{proj}$, we can obtain the projected feature ($P = [p_1,...,p_T]$), which is represented by
\begin{equation}
\begin{aligned}
& \label{fp}
p_t = F_{proj}(F_{tcn}(s_t)) = F_{proj}(z_t). 
\end{aligned}%
\end{equation}
Following \cite{kalantidis2020hard, khosla2020supervised, wang2021exploring}, the training samples' selection for our sample-level contrastive loss is also crucial, where the harder positives and negatives would bring more contributions to the model's optimization. Therefore, we adopt a Hybrid Hard Example Sampling strategy for our task. In the supervised setting, the sample's hardness depends on whether it is classified correctly or not \cite{khosla2020supervised}. Therefore, in each class, we first randomly choose the samples' projected features with incorrect predictions as half of the example samples. If the number of incorrect predictions is less than our preset value, we additionally select samples at the activity boundary as a supplement. Then, the rest of the example samples are randomly sampled in the wearable sequence. This sampling strategy has been shown effective in supervised contrastive learning works \cite{khosla2020supervised, wang2021exploring}, and for our specific task, we supplemented samples near the activity boundary as harder samples.
Therefore, we employ the supervised contrastive loss on these example samples ($\mathcal{I}$) to get the sample-level contrastive loss:
\begin{align}
\label{slc}
\resizebox{.85\linewidth}{!}{$
\displaystyle
L_{SC} = \frac{1}{\left| \mathcal{I} \right|} \sum\limits_{p_i\in \mathcal{I}} \frac{1}{\left| \mathcal{P}_i \right|} \sum\limits_{p_j\in \mathcal{P}_i} \frac{\exp{(p_i \cdot p_j / \tau)}}{\exp{(p_i \cdot p_j / \tau)} + \sum\limits_{p_n\in \mathcal{N}_i} \exp{(p_i \cdot p_n / \tau)}}
$},
\end{align}%
where $\mathcal{P}_i$ and $\mathcal{N}_i$ are the projected feature sets of the positive and negative samples in our method's example samples ($\mathcal{I}$). 

It is noted that the feature representation learned based on a single sample is not efficient. Besides, only using sample-level loss may make the model concentrate on the individual sample and ignore the feature information of the whole classes within a segment. Therefore, we integrate the sample-level features with the same activity class to obtain segment-level features based on the activity ground-truth labels. Then, the segment-level projected features would augment the original example samples ($\mathcal{I}$), which could form the sample-level (sample-to-sample) and segment-level contrast (sample-to-segment and segment-to-segment) with Eq.(\ref{slc}) on the new samples set $\mathcal{\widetilde{I}}$:
\begin{align}
\label{alc}
\resizebox{.85\linewidth}{!}{$
\displaystyle
L_{CON} = \frac{1}{\left| \mathcal{\widetilde{I}} \right|} \sum\limits_{p_i\in \mathcal{\widetilde{I}}} \frac{1}{\left| \mathcal{\widetilde{P}}_i \right|} \sum\limits_{p_j\in \mathcal{\widetilde{P}}_i} \frac{\exp{(p_i \cdot p_j / \tau)}}{\exp{(p_i \cdot p_j / \tau)} + \sum\limits_{p_n\in \mathcal{\widetilde{N}}_i} \exp{(p_i \cdot p_n / \tau)}}
$},
\end{align}%
where $\mathcal{\widetilde{P}}_i$ and $\mathcal{\widetilde{N}}_i$ are the positive and negative samples sets after supplementing segment-level features.

With the sample-level and segment-level contrast, the multi-level contrastive loss could regularize the embedding space for better performance. Therefore, based on the multi-level contrast-based regularization, our model can be optimized by:
\begin{equation}
\begin{aligned}
\label{eq1}
\min_{\theta_{t},\theta_{c},\theta_p} L_C + \lambda L_{CON},
\end{aligned}
\end{equation}%
where $ \theta_{t},\theta_{c},\theta_p $ are the parameters of $F_{tcn}, F_{cls}, F_{proj}$ and $\lambda$ is the hyperparameter of the multi-level contrastive loss.
\begin{table*}[]
	\centering
	\caption{\sc The Segmentation and Recognition Performance on PAMAP2 and Hospital Datasets}
	\begin{tabular}{c|cccc|cccc}
		\toprule
		& \multicolumn{4}{c|}{PAMAP2}     &\multicolumn{4}{c}{Hospital}                          \\ \cmidrule{2-5} \cmidrule{6-9} 
		& $Precision$ & $Recall$ & $F_m$  & $JI$  & $Precision$ & $Recall$  & $F_m$  & $JI$     \\ \midrule
		DeepConvLSTM\cite{ordonez2016deep}    & 84.6287\%      & 86.1885\%        & 83.9823\%     & 74.4746\%              & 65.6113\%    & 62.2837\%       & 62.8305\%    & 52.5231\%           \\
		Attend\&Discriminate\cite{abedin2021attend} &91.7781\%     & 90.4745\%  & 90.8253\%     & 84.1163\%              & 71.0255\%  & 63.9932\%       & 66.6240\%  & 55.5557\%        \\
		Dense Labeling-FCN\cite{yao2018efficient}   & 83.4224\%   & 87.8783\%   & 83.8004\%   & 75.4087\%                   & 69.7513\%    & 67.8851\%   & 68.4273\%    & 57.7744\%          \\
		Dense Labeling-TCN\cite{farha2019ms}    & 92.0987\%     & 90.1461\%   & 90.4185\%     & 83.5553\%              & 77.6318\%     & 77.4675\%       & 77.3319\%     & 66.2596\%         \\  \midrule
		Ours   & \textbf{93.2814\%} & \textbf{93.4794\%}     & \textbf{93.1156\%} & \textbf{87.6325\%}  & \textbf{81.0295\%} & \textbf{84.4179\%}  & \textbf{82.4158\%} & \textbf{72.0883\%} \\\bottomrule
	\end{tabular}
	\label{all}
	\vspace{-0.5cm}
\end{table*}
\subsection{Multi-stage Architecture for Wearables-based Activity Segmentation and Recognition}
Recent works \cite{wei2016convolutional, farha2019ms, xia2022boundary} have shown that multi-stage structure can improve the performance of various tasks. The core idea behind the multi-stage structures is that the current stage network takes the previous stage's predictions as input and gets the refined result as output. Therefore, to address the challenge that the fine-grained information in predictions might be discarded in the single-stage TCN, we introduce the MS-TCN for our wearable-based human activity segmentation and recognition. 

As shown in Fig. \ref{Overview}, we stack multiple SS-TCN architectures, where each stage would take the sample-level predictions from the previous SS-TCN stage's output as inputs. In the training phase, in addition to using the sample-wise cross-entropy loss to constrain the optimization of SS-TCN parameters, the multi-level contrastive learning loss is also employed at each stage. 

Therefore, with the above technology introduction, we can optimize our proposed model by the final objective function:
\begin{equation}
\begin{aligned}
\label{alleq}
\min_{\theta_{T},\theta_{C},\theta_P} \sum_n^{N_s} (L_C^n + \lambda L_{CON}^n),
\end{aligned}
\end{equation}%
where $N_s$ is the stage number, and $ \theta_{T}=\{\theta_{t}^1,...,\theta_{t}^{N_s}\}$, $\theta_{C}=\{\theta_{c}^1,...,\theta_{c}^{N_s}\}$, $\theta_P=\{\theta_{p}^1,...,\theta_{p}^{N_s}\}$ are our model's parameters.

\section{Experiment}
\label{exp}
\subsection{Experimental Setup}
In our experiment, we choose the PAMAP2 \cite{reiss2012introducing} and Hospital \cite{yao2018efficient} dataset for evaluation. And our evaluation protocol is kept same with \cite{reiss2012introducing} and \cite{yao2018efficient} and we use the Jaccard Index ($JI$) \cite{gammulle2021tmmf}, Class-average F1-score ($F_m$) \cite{abedin2021attend}, Precision and Recall \cite{pei2020mars} to comprehensively illustrate the competing methods' performance. We trained our model on a PC with a GPU RTX 3090, where the batch size and learning rate are 32 and 0.001, respectively. The hyperparameter $\lambda$ is set as 1.0, and we set the stage number to 2 and 3 on the PAMAP2 and Hospital datasets.

This following will introduce the selected datasets in our experiment:
\begin{itemize}
\item \textbf{Hospital} \cite{yao2018efficient} dataset is a continuous activity sequence collected in a hospital with 12 subjects using a single IMU sensor (including acceleration and angular velocity). This dataset contains 7 daily activities, which include basic activities and transition activities, such as sitting and sitting down. In our experiment, we adopt the evaluation strategy with the same settings as \cite{yao2018efficient}.

\item \textbf{PAMAP2} \cite{reiss2012introducing} dataset is a classic sensor-based HAR dataset, which is widely used in the evaluation of various HAR algorithms. We selected 12 types of activities in this dataset for this experiment, in which there are relatively few activity transitions, but some high-level semantic activities, such as ironing, etc. Following \cite{abedin2021attend}, our experiment use the Leave-One-Subject-Out (LOSO) evaluation strategy, where Participant 5 and 6 are set as the validation-set and test-set, respectively.

\end{itemize}

	
\begin{table}[]
	\centering
	\caption{\sc  The Ablation Study on Two Public Datasets}
	\resizebox{0.5\textwidth}{!}{ 

		\begin{tabular}{c|ccc|cc|cc}
			\toprule
             \multicolumn{1}{c|}{\multirow{3}{*}{No.}} & \multicolumn{3}{c|}{Modules}    & \multicolumn{4}{c}{Datasets}   \\ \cmidrule{2-8} 
			\multicolumn{1}{c|}{}                     & \multicolumn{1}{c}{\multirow{2}{*}{\begin{tabular}[c]{@{}c@{}}Multi-\\Stage\\  Form\end{tabular}}} & \multicolumn{1}{c}{\multirow{2}{*}{\begin{tabular}[c]{@{}c@{}}Sample-\\level\\  Contrast\end{tabular}}} & \multicolumn{1}{c}{\multirow{2}{*}{\begin{tabular}[c]{@{}c@{}}Segment-\\level\\  Contrast\end{tabular}}} & \multicolumn{2}{|c|}{PAMAP2}                                  & \multicolumn{2}{c}{Hospital}   \\ \cmidrule{5-6}  \cmidrule{7-8}
			\multicolumn{1}{c|}{}                     & \multicolumn{1}{c}{}                                                                         & \multicolumn{1}{c}{}                                                                     & \multicolumn{1}{c}{}                                                                      & \multicolumn{1}{|c}{$F_m$} & \multicolumn{1}{c}{$JI$} & \multicolumn{1}{|c}{$F_m$} & \multicolumn{1}{c}{$JI$}  \\
			\midrule
			(1)                                       & \XSolidBrush                                                                                             & \XSolidBrush                                                                                         &\XSolidBrush                                                                                           &90.42\%                              &83.56\%                              &77.33\%             &66.26\%        \\
			(2)                                       &     \CheckmarkBold                                                                                        &   \XSolidBrush             &        \XSolidBrush                                 &91.27\%        &85.11\%         &79.24\%                             &  67.90\%   \\
			(3)  &  \XSolidBrush   &       \CheckmarkBold    &     \CheckmarkBold     &92.20\%     &86.60\%       &78.28\%       &67.18\%    \\
			(4)    &  \CheckmarkBold  &   \CheckmarkBold    & \XSolidBrush     &92.36\%          &86.83\%                          &80.52\%    &69.84\%  \\
			(5)   &  \CheckmarkBold & \CheckmarkBold  &  \CheckmarkBold     &\textbf{93.12\%}   &\textbf{87.63\% }  &\textbf{82.42\%}  &\textbf{72.09\% } \\

			\bottomrule 
	\end{tabular}}
\label{ablation}
\vspace{-0.5cm}
\end{table}

\subsection{Overall Experiment Results}
\subsubsection{Overall Segmentation and Recognition Performance Comparison}
We compare our proposed method with various state-of-the-art approaches: (1) DeepConvLSTM \cite{ordonez2016deep}; (2) Attend\&Discriminate \cite{abedin2021attend}; (3) Dense Labeling-FCN \cite{yao2018efficient}; (4) Dense Labeling-TCN \cite{farha2019ms}. Methods (1) and (2) are traditional methods with fixed-sized sliding windows, where we set the window size to 24 and the stride to 1 for predicting each sample's label. Methods (3) and (4) are the joint activity segmentation and recognition methods, which directly predict the sample-level activity labels. Table \ref{all} shows the segmentation and recognition performance of these methods, where we can summarize the experimental phenomena as follow:

(1) Our proposed method outperforms the state-of-the-art algorithms with segmentation and recognition metrics on these two public datasets. For the recognition performance measured by class-average F1-score, our proposed approach is 2.2903\%/5.8039\% higher than Attend\&Discriminate/Dense Labeling-TCN on the PAMAP2/Hospital datasets, respectively. Moreover, in terms of segmentation performance ($JI$), our method still achieves promising results on both datasets, which are improved by 3.5162\% and 5.8287\% at least, reaching 87.6325\% and 72.0883\%.

(2) The Dense Labeling-FCN \cite{yao2018efficient} and TCN \cite{farha2019ms} are the sample-level activity prediction methods. For the Hospital dataset containing many activity transitions, both the segmentation and recognition performance of the sample-level activity prediction methods are better than the traditional wearable-based HAR methods with fixed-size sliding windows.

Moreover, more comparative analysis will be introduced further in Section \ref{exp}-B
\begin{figure*}[ht]
	\centering
	\subfigure[Attend\&Discriminate]{\includegraphics[width=5.4cm]{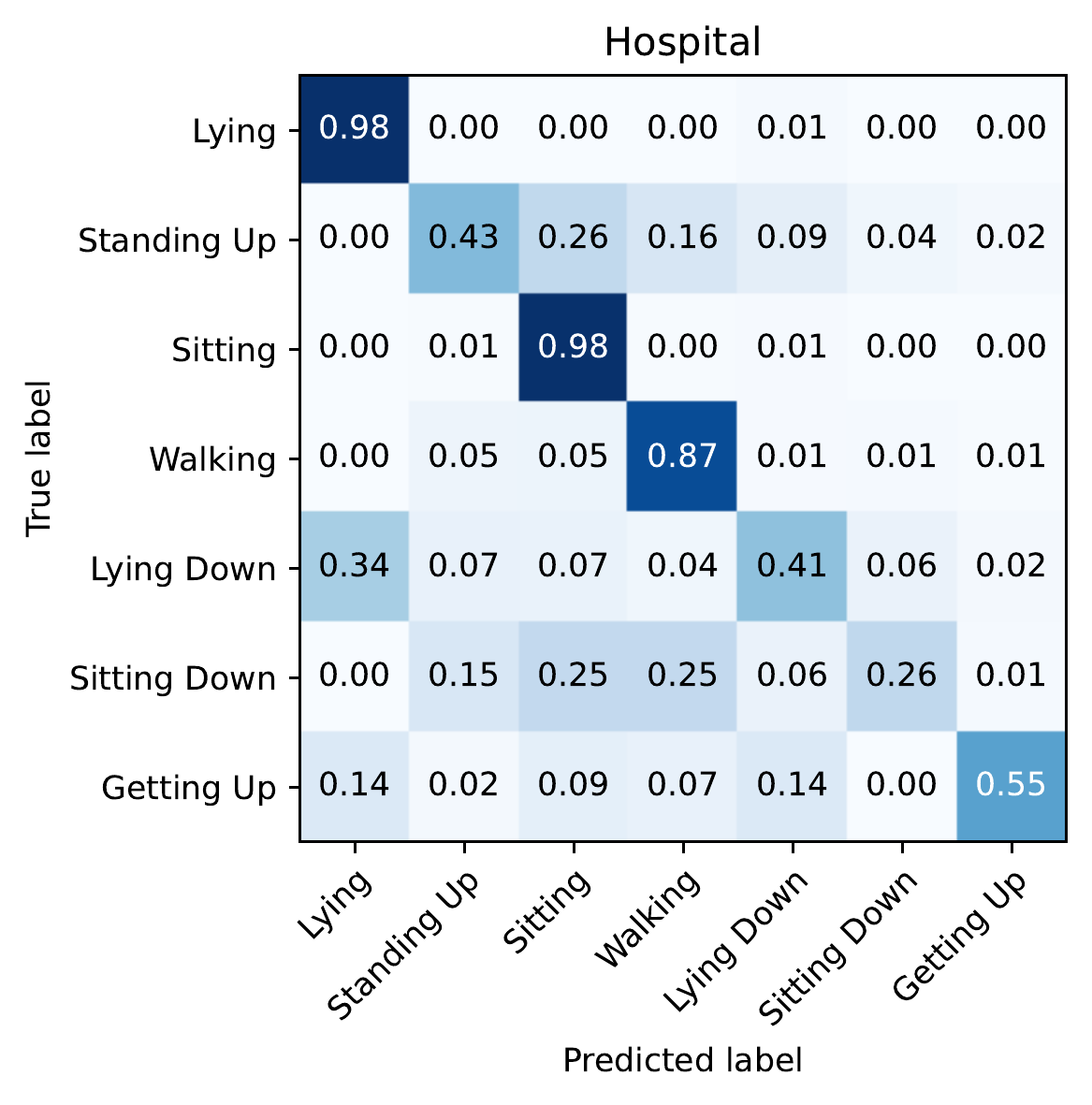}}
	\label{ADM}
	\subfigure[Dense Labeling-FCN ]{\includegraphics[width=5.4cm]{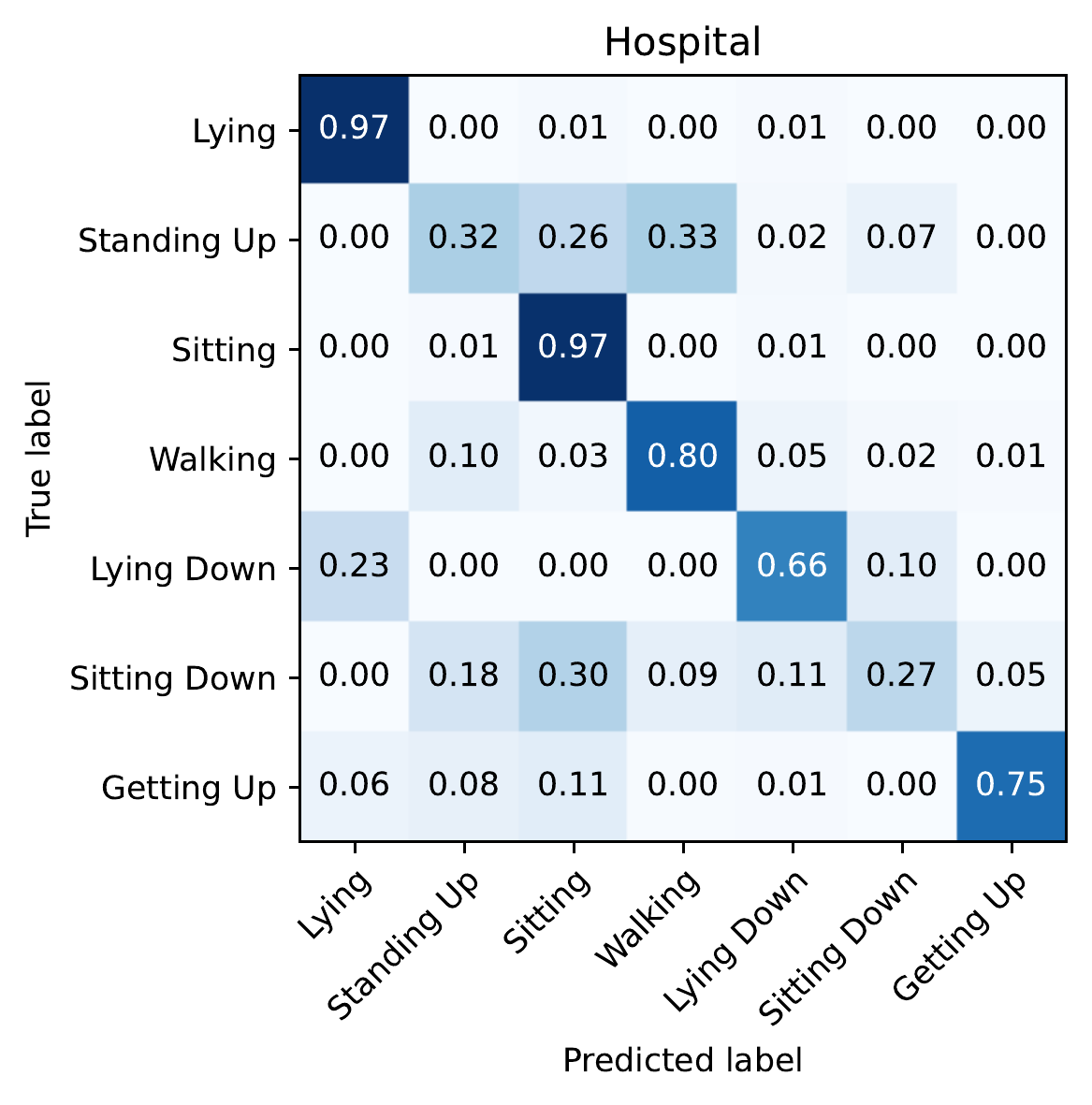}}
	\label{FCNM}  
	\subfigure[Ours]{\includegraphics[width=5.4cm]{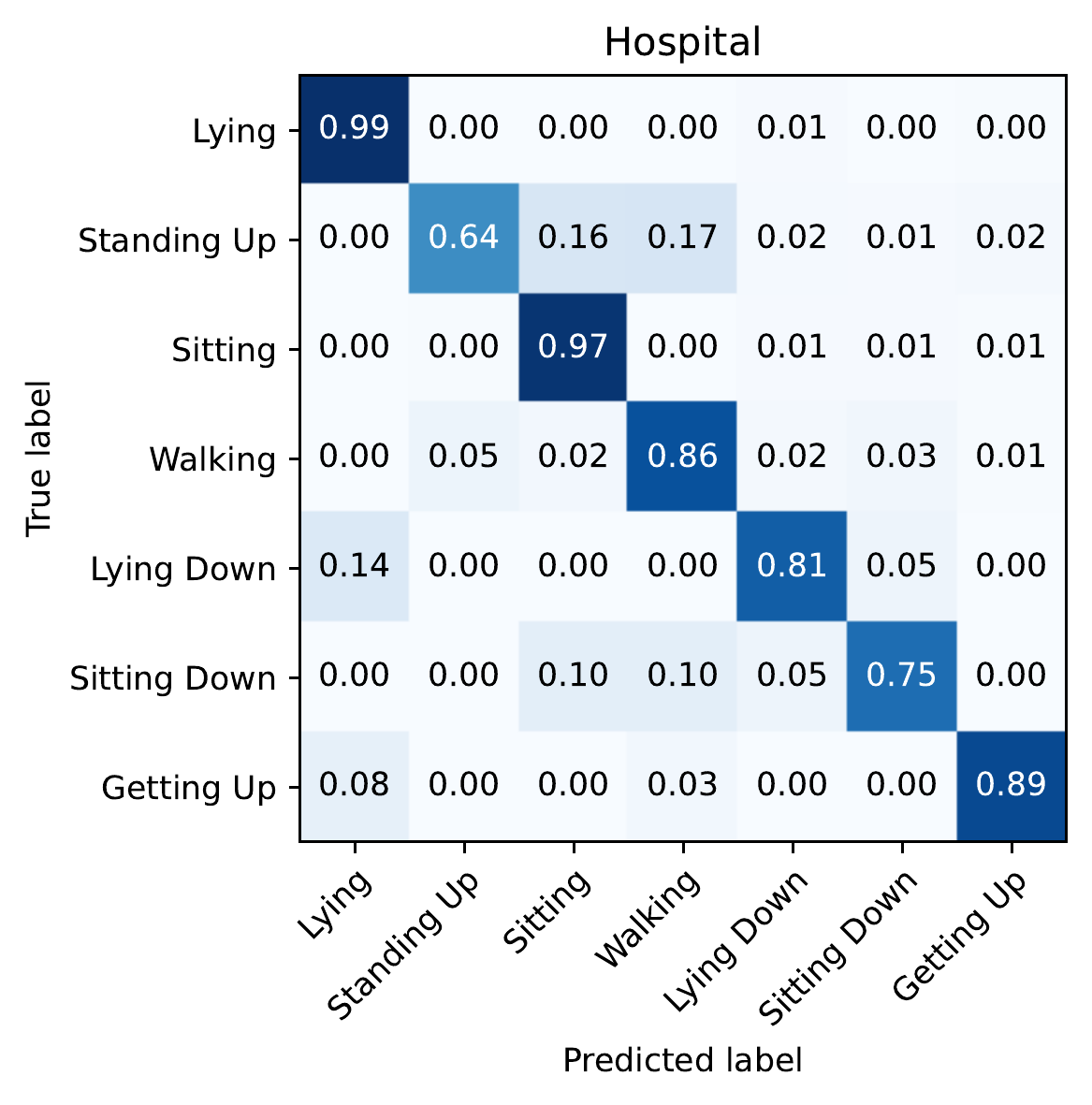}}
	\label{OURM} 
	\caption{The confusion matrix of (a) Attend\&Discriminate (b) Dense Labeling-FCN (c) Ours methods on Hospital dataset.}
	\label{Matrics}
	\vspace{-0.5cm}
\end{figure*}.
\begin{figure}[t] 
	\subfigure[Hospital]{
		\includegraphics[width=8cm]{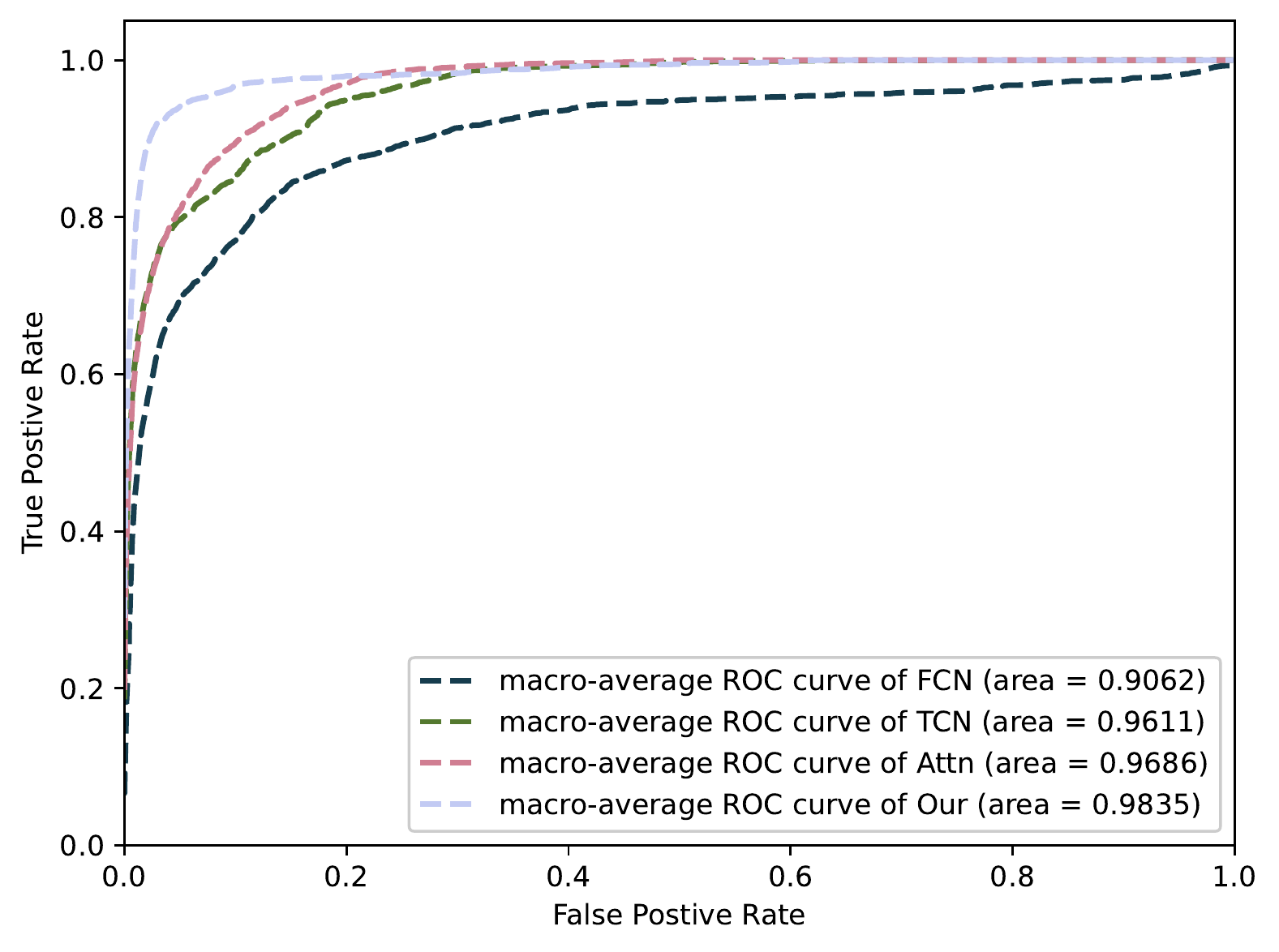}} 
	\subfigure[PAMAP2]{
		\includegraphics[width=8cm]{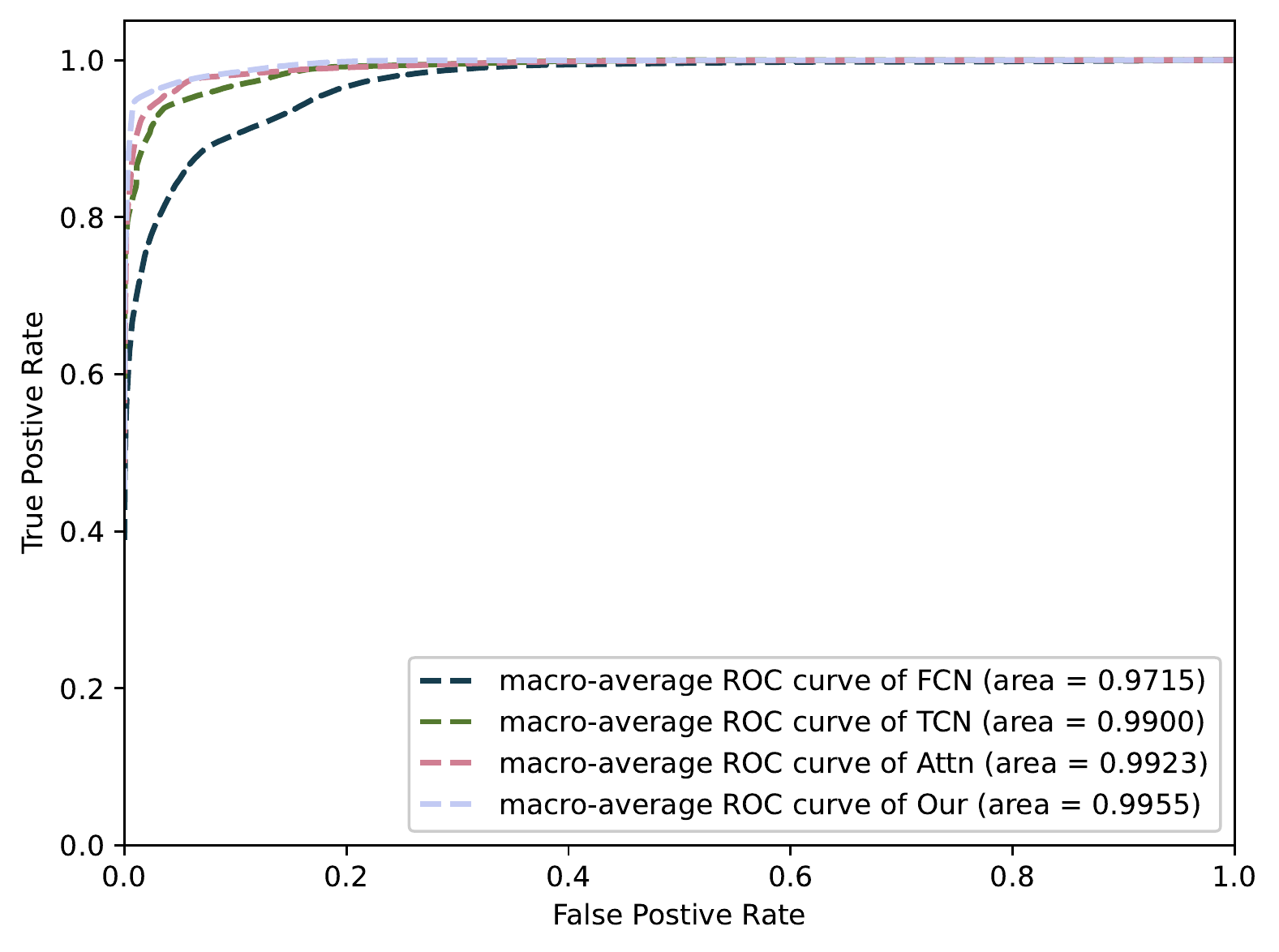}} 
	\caption{The ROC curve of Hospital and PAMAP2 datasets.}
	\label{ROC}
	\vspace{-0.5cm}
\end{figure}
\subsubsection{Ablation Study}
Furthermore, an ablation study is summarized in Table \ref{ablation}, which shows the designed modules' effectiveness in our proposed approach. Therefore, we consider five variants: \textit{(1)} the baseline with single-stage TCN and sample-wise cross-entropy loss; \textit{(2)} a multi-stage TCN network with sample-wise cross-entropy loss; \textit{(3)} the baseline \textit{(1)} with sample- and segment-level contrast modules; \textit{(4)} the multi-stage TCN network only with the sample-level contrast module; \textit{(5)} our final proposed model with multi-stage architecture and multi-level contrast modules.
\begin{figure}[t] 
	\subfigure[Hospital]{
		\includegraphics[width=0.45\textwidth]{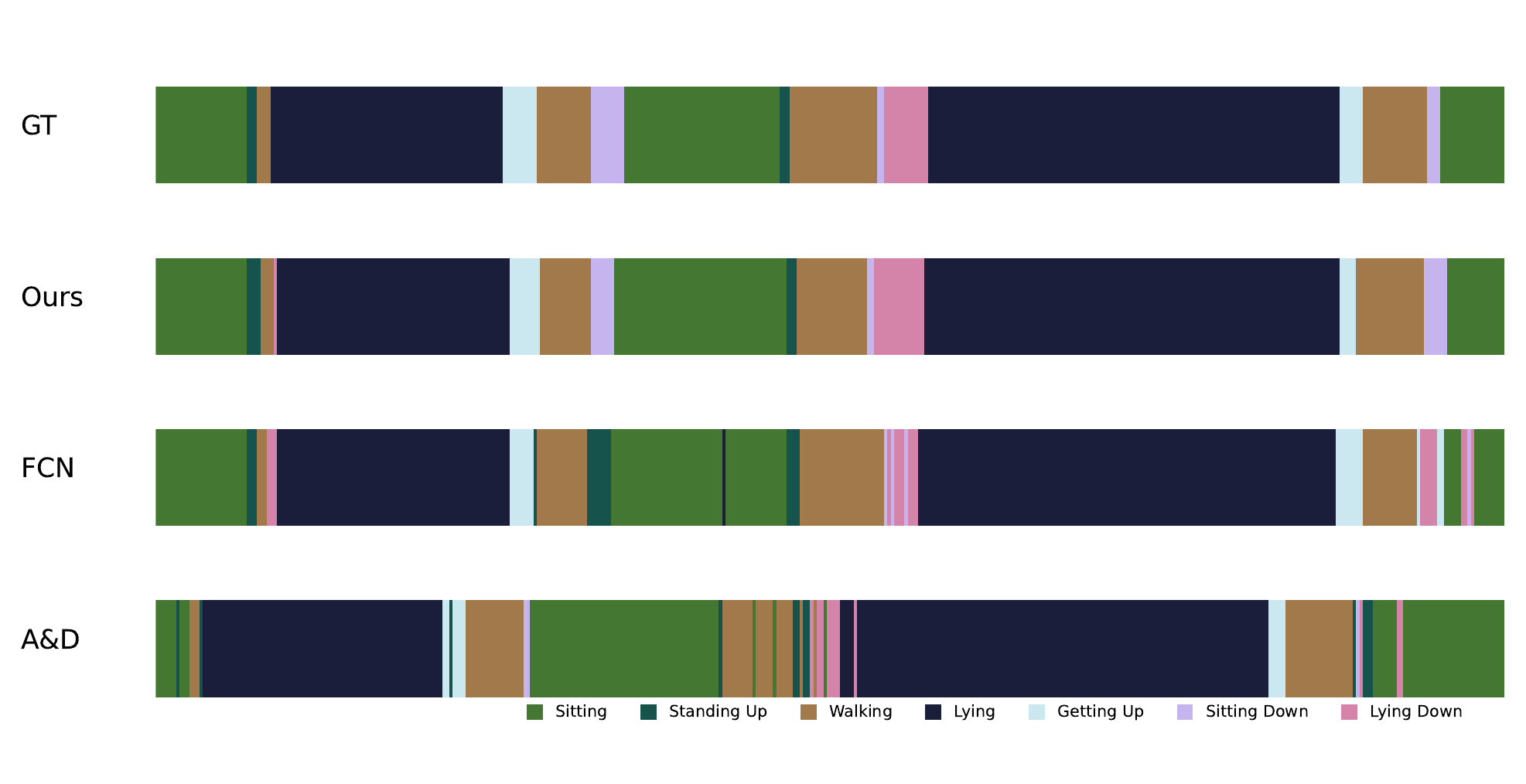}} \\ 
	\subfigure[PAMAP2]{
		\includegraphics[width=0.45\textwidth]{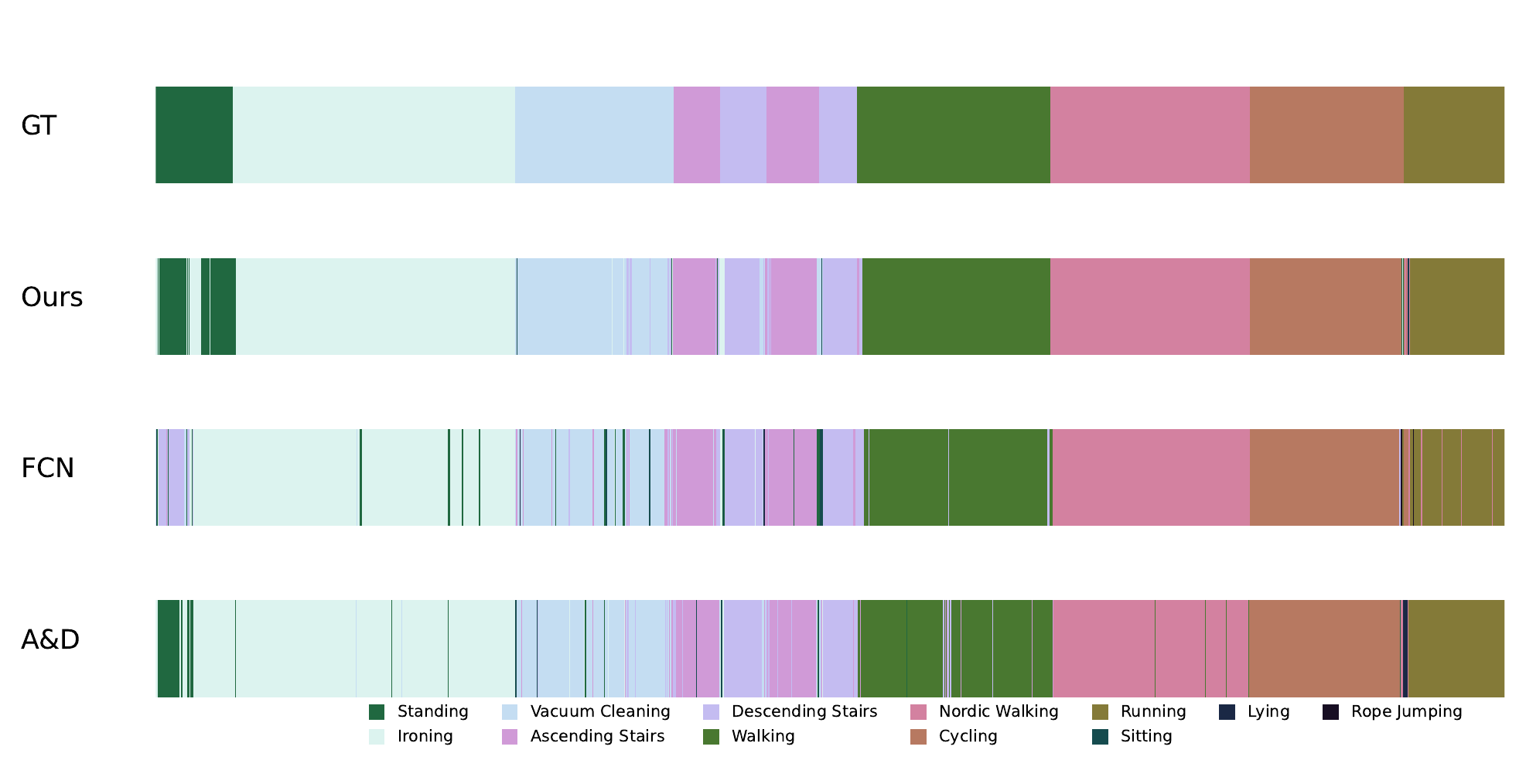}} 
	\caption{Qualitative results visualization. (a), (b) in comparison with the competing method on Hospital and PAMAP2 datasets.}
	\label{DIngxing}
	\vspace{-0.5cm}
\end{figure}

\begin{figure*}[t]
	\centering
	\subfigure[Dense Labeling-FCN]{\includegraphics[width=5.9cm]{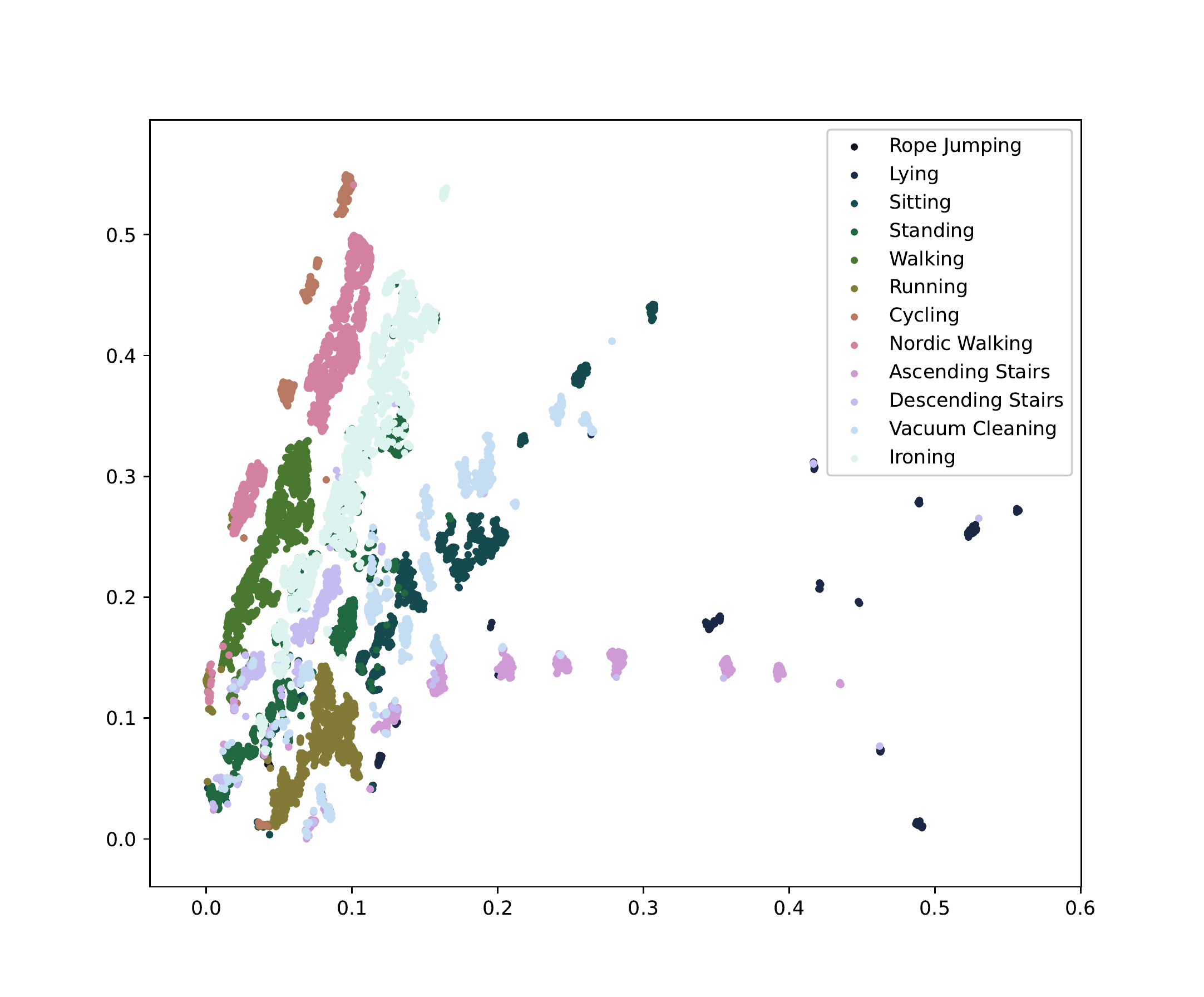}}
	\label{O-t-Yuan}
	\subfigure[Dense Labeling-TCN]{\includegraphics[width=5.9cm]{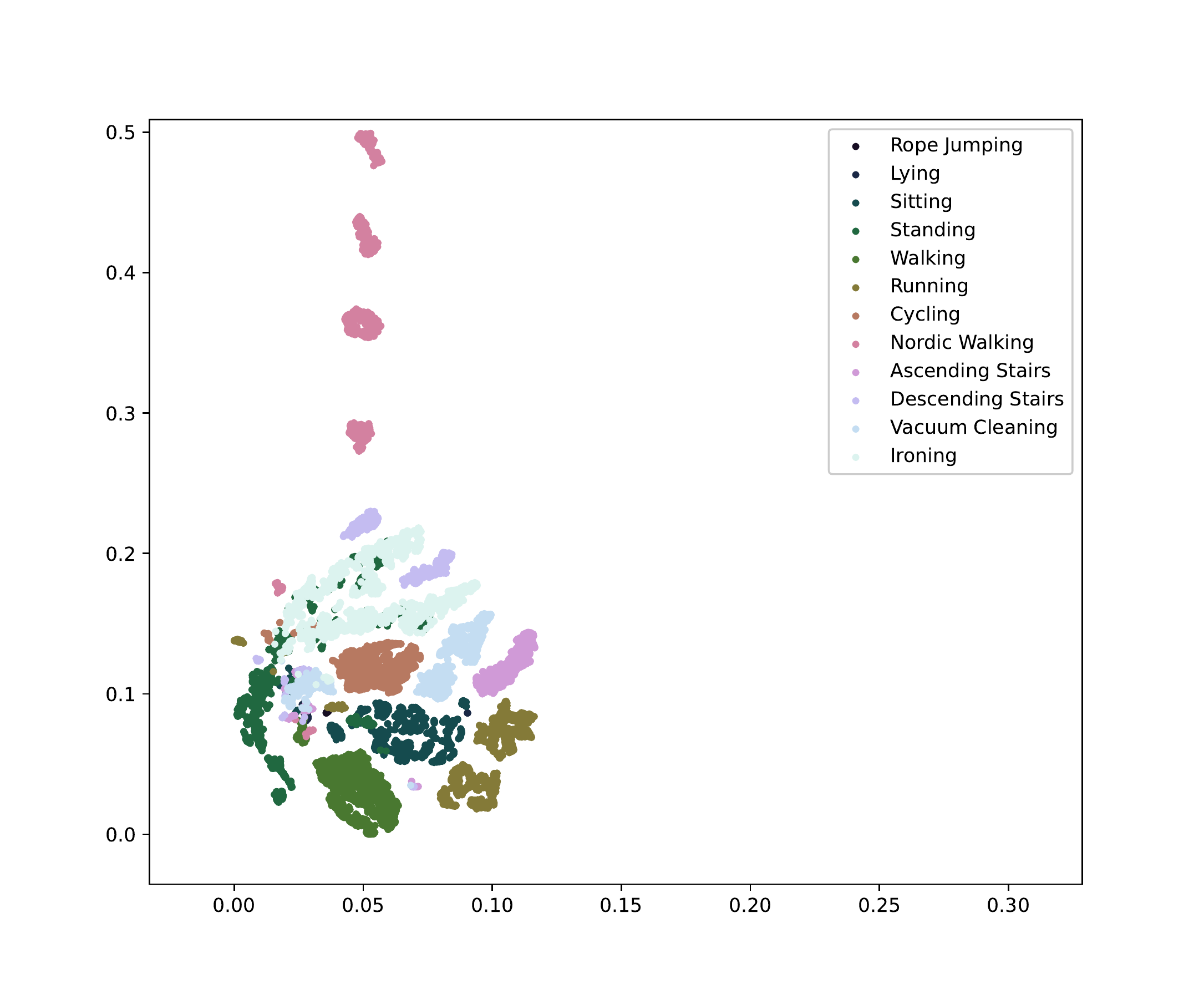}}
	\label{O-t-ms}
	\subfigure[Ours]{\includegraphics[width=5.9cm]{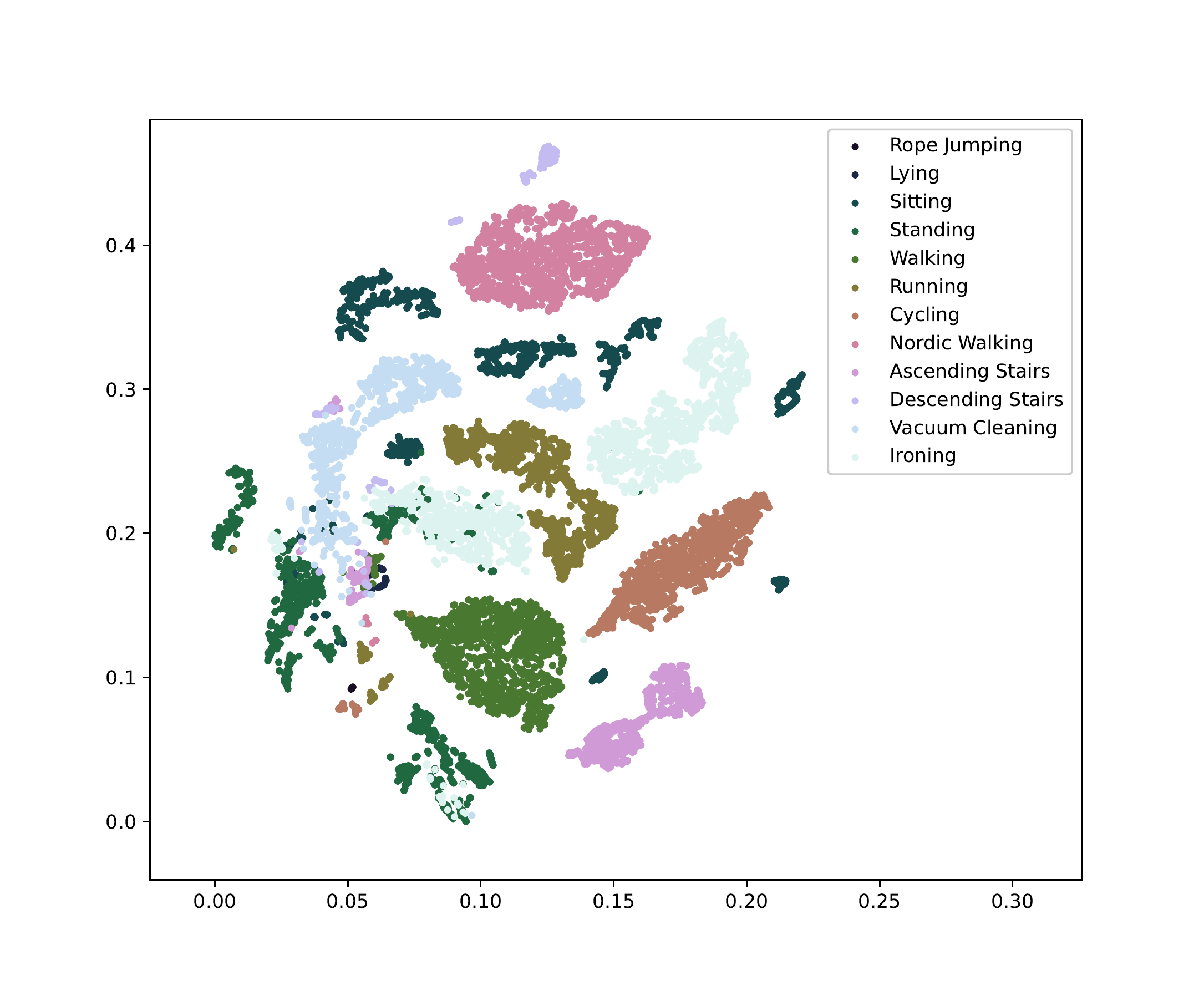}}
	\label{O-t-our}
	\caption{The t-SNE visualization of different methods on PAMAP2 dataset. (a) The t-SNE visualization of Dense Labeling-FCN learned representation. (b) The t-SNE visualization of Dense Labeling-TCN learned representation. (c) The t-SNE visualization of our approach learned representation.}
	\label{Opp-tsne}
\end{figure*}
As shown in Tab. \ref{ablation}, the proposed approach with all customized modules \textit{(5)} obtains the best segmentation and recognition performance, where 5.08\%/5.83\% improvements have been obtained on the Hospital dataset with $F_m$ and $JI$. By comparing the performance differences between \textit{(1)} and \textit{(2)} and the ones between \textit{(3)} and \textit{(5)}, we can find that the multi-stage structure has a positive impact on improving the segmentation and recognition performance. For evaluating the effectiveness of the contrast module, the performance of \textit{(2)} and \textit{(5)} are better than those without contrast module (as in \textit{(1)} and \textit{(2)}), respectively. Moreover, for \textit{(2)}, \textit{(4)}, \textit{(5)}, we further demonstrate the utility of multi-level (sample-level and segment-level) contrastive learning modules. The performance of variant \textit{(4)} only with the sample-level contrast module outperforms the multi-stage baseline \textit{(2)}. With the segment-level contrast, the overall design \textit{(5)} achieves further improvement in segmentation and recognition performance, which are increased by 0.76\% and 0.81\% compared with \textit{(2)} on the PAMAP2 dataset. Therefore, with the ablation study, the designed modules in our proposed approach effectively promote wearable-based HAR task performance improvement.

\subsection{Analysis}
To further demonstrate the performance of our method, we will illustrate four aspects: confusion matrix, ROC (Receiver Operating Characteristic) curve, qualitative evaluation and the embedding visualization with t-SNE.

\subsubsection{Analysis on the Confusion Matrix}
The confusion matrices of Attend\&Discriminate, Dense Labeling-FCN and our proposed approach are shown in Fig. \ref{Matrics} (a), (b) and (c). Comparing the results in the confusion matrices, we can find that the most classes have higher recognition accuracy in our proposed approach and the transition activities ("Standing Up", "Lying Down", "Sitting Down" and "Getting Up") are more challenging than the quasi-static activities ("Lying", "Sitting" and "Walking"). For the transition activities, the sample-level activity prediction methods always achieve better recognition performance than the traditional methods with the fixed-length sliding window. Furthermore, our approach has significant improvements in these transition activities, which are increased by 21\%/32\% (Standing Up), 40\%/15\% (Lying Down), 49\%/48\% (Sitting Down) and 34\%/14\% (Getting Up) for the Attend\&Discriminate and Dense Labeling-FCN, respectively, demonstrating of the superiority of joint segmentation and recognition. 

\subsubsection{Analysis on ROC Curve}
For a fair evaluation of the classifiers, the Receiver Operating Characteristic (ROC) and Area under the Curve (AUC) are adopted. The horizontal and vertical axes of the ROC curve are the false positive rate and the true positive rate, where the curve value is varied with the discrimination threshold \cite{hanley1983method}. Fig. \ref{ROC} shows the effectiveness of the proposed method, achieving the surpassing performance of 0.9835/0.9955 on the Hospital/PAMAP2 datasets.

\subsubsection{Qualitative Evaluation}
Fig. \ref{DIngxing} shows qualitative results using different colors for different activity classes, where (a) represents the fragment with samples 6000-10000 on the Hospital dataset and (b) shows the qualitative results with 20000-100000 samples of the PAMAP2 dataset. The Attend\&Discriminate (denoted by "A\&D") and Dense Labeling-FCN (denoted by "FCN") were employed for a fair comparison. Compared with sample-level activity ground-truth labels (notation "GT"), our method (notation "Ours") attains the best performance. This can be explained by the novel designs in the proposed method, reducing the fluctuations in predictions and enhancing sensitiveness to the short-term and transition activities.

\subsubsection{Feature Embedding Visualization with t-SNE}
The t-distributed Stochastic Neighbor Embedding (t-SNE) plots have been utilized to illustrate the structured embedded spaces. Due to the limit space, Fig. \ref{Opp-tsne} shows the t-SNE result on only the PAMAP2 dataset. With the embedding space visualizations, we can find that our proposed approach has a more structured embedded representation for each activity, demonstrating the superior recognition performance of the proposed method again.

\section{Conclusion}
This paper first innovatively introduced joint activity segmentation and recognition framework to deal with the challenging multi-class window problem and alleviate the vagueness issues in HAR. We then proposed a multi-stage sample-level activity prediction model with a multi-level contrast module, extracting the effective representation and learning a well-structured embedding space. Extensive experimental results on the two public HAR datasets demonstrated that our proposed approach obtained the prevailing segmentation and recognition performance, reaching 87.63\% ($JI$)/93.12\% ($F_m$) and 72.09\% ($JI$)/82.42\% ($F_m$) on the PAMAP2 and Hospital datasets. Finally, we will provide the efficiency analysis and extend the proposed method to the HAR framework with zero-/few- shot samples for our future work. 
\section*{Acknowledgment}
Dr. Pei's work was supported by the National Natural Science Foundation of China (Grant No.61873163), partly by the Shanghai Science and Technology Committee under Grant 20511103103. Dr. Chu's work was supported by the International Postdoctoral Exchange Fellowship.

\bibliography{ref}

\begin{thebibliography}{10}
\providecommand{\url}[1]{#1}
\csname url@samestyle\endcsname
\providecommand{\newblock}{\relax}
\providecommand{\bibinfo}[2]{#2}
\providecommand{\BIBentrySTDinterwordspacing}{\spaceskip=0pt\relax}
\providecommand{\BIBentryALTinterwordstretchfactor}{4}
\providecommand{\BIBentryALTinterwordspacing}{\spaceskip=\fontdimen2\font plus
\BIBentryALTinterwordstretchfactor\fontdimen3\font minus
  \fontdimen4\font\relax}
\providecommand{\BIBforeignlanguage}[2]{{%
\expandafter\ifx\csname l@#1\endcsname\relax
\typeout{** WARNING: IEEEtran.bst: No hyphenation pattern has been}%
\typeout{** loaded for the language `#1'. Using the pattern for}%
\typeout{** the default language instead.}%
\else
\language=\csname l@#1\endcsname
\fi
#2}}
\providecommand{\BIBdecl}{\relax}
\BIBdecl

\bibitem{chen2020creativebioman}
M.~Chen, Y.~Jiang, Y.~Cao, and A.~Y. Zomaya, ``Creativebioman: a brain-and
  body-wearable, computing-based, creative gaming system,'' \emph{IEEE Trans.
  Syst. Man Cybern. Syst.}, vol.~6, no.~1, pp. 14--22, 2020.

\bibitem{gammulle2021tmmf}
H.~Gammulle, S.~Denman, S.~Sridharan, and C.~Fookes, ``Tmmf: Temporal
  multi-modal fusion for single-stage continuous gesture recognition,''
  \emph{IEEE Trans. Image Process.}, vol.~30, pp. 7689--7701, 2021.

\bibitem{ahuja2021vid2doppler}
K.~Ahuja, Y.~Jiang, M.~Goel, and et~al., ``Vid2doppler: Synthesizing doppler
  radar data from videos for training privacy-preserving activity
  recognition,'' in \emph{Int. Conf. Comput. Hum. Interact.}, 2021, pp. 1--10.

\bibitem{pei2020mars}
L.~Pei, S.~Xia, L.~Chu, and et~al., ``Mars: Mixed virtual and real wearable
  sensors for human activity recognition with multi-domain deep learning
  model,'' \emph{IEEE Internet Things J.}, pp. 1--1, 2021.

\bibitem{zhang2021open}
Z.~Zhang, L.~Chu, S.~Xia, and L.~Pei, ``Open set mixed-reality human activity
  recognition,'' in \emph{IEEE Glob. Commun. Conf. (GLOBECOM)}.\hskip 1em plus
  0.5em minus 0.4em\relax IEEE, 2021, pp. 1--7.

\bibitem{chu2021ahed}
L.~Chu, L.~Pei, and R.~Qiu, ``Ahed: A heterogeneous-domain deep learning model
  for iot-enabled smart health with few-labeled eeg data,'' \emph{IEEE Internet
  Things J.}, vol.~8, no.~23, pp. 16\,787--16\,800, 2021.

\bibitem{chen2017wearable}
M.~Chen, Y.~Ma, Y.~Li, D.~Wu, Y.~Zhang, and C.-H. Youn, ``Wearable 2.0:
  Enabling human-cloud integration in next generation healthcare systems,''
  \emph{IEEE Commun. Mag.}, vol.~55, no.~1, pp. 54--61, 2017.

\bibitem{ma2019attnsense}
H.~Ma, W.~Li, X.~Zhang, S.~Gao, and S.~Lu, ``Attnsense: Multi-level attention
  mechanism for multimodal human activity recognition.'' in \emph{Int. Joint
  Conf. Artif. Intell. (IJCAI)}, 2019, pp. 3109--3115.

\bibitem{chen2021deep}
K.~Chen, D.~Zhang, L.~Yao, B.~Guo, Z.~Yu, and Y.~Liu, ``Deep learning for
  sensor-based human activity recognition: Overview, challenges, and
  opportunities,'' \emph{ACM Comput. Surv.}, vol.~54, no.~4, pp. 1--40, 2021.

\bibitem{xia2021learning}
S.~Xia, L.~Chu, L.~Pei, and et~al., ``Learning disentangled representation for
  mixed-reality human activity recognition with a single imu sensor,''
  \emph{IEEE Trans. Instrum. Meas.}, vol.~70, pp. 1--14, 2021.

\bibitem{radu2018multimodal}
V.~Radu, C.~Tong, S.~Bhattacharya, N.~D. Lane, C.~Mascolo, M.~K. Marina, and
  F.~Kawsar, ``Multimodal deep learning for activity and context recognition,''
  \emph{Proc. ACM Interact. Mob. Wearable Ubiquitous Technol. (IMWUT)}, vol.~1,
  no.~4, pp. 1--27, 2018.

\bibitem{yang2015deep}
J.~Yang, M.~N. Nguyen, P.~P. San, X.~Li, and S.~Krishnaswamy, ``Deep
  convolutional neural networks on multichannel time series for human activity
  recognition.'' in \emph{Int. Joint Conf. Artif. Intell. (IJCAI)},
  vol.~15.\hskip 1em plus 0.5em minus 0.4em\relax Buenos Aires, Argentina,
  2015, pp. 3995--4001.

\bibitem{guan2017ensembles}
Y.~Guan and T.~Pl{\"o}tz, ``Ensembles of deep lstm learners for activity
  recognition using wearables,'' \emph{Proc. ACM Interact. Mob. Wearable
  Ubiquitous Technol. (IMWUT)}, vol.~1, no.~2, pp. 1--28, 2017.

\bibitem{ordonez2016deep}
F.~J. Ord{\'o}{\~n}ez and D.~Roggen, ``Deep convolutional and lstm recurrent
  neural networks for multimodal wearable activity recognition,''
  \emph{Sensors}, vol.~16, no.~1, p. 115, 2016.

\bibitem{yao2018efficient}
R.~Yao, G.~Lin, Q.~Shi, and D.~C. Ranasinghe, ``Efficient dense labelling of
  human activity sequences from wearables using fully convolutional networks,''
  \emph{Pattern Recognit.}, vol.~78, pp. 252--266, 2018.

\bibitem{xia2022boundary}
S.~Xia, L.~Chu, L.~Pei, W.~Yu, and R.~Qiu, ``A boundary consistency-aware
  multi-task learning framework for joint activity segmentation and recognition
  with wearable sensors,'' \emph{IEEE Trans. Ind. Informat.}, 2022.

\bibitem{long2015fully}
J.~Long, E.~Shelhamer, and T.~Darrell, ``Fully convolutional networks for
  semantic segmentation,'' in \emph{Proc. IEEE Comput. Soc. Conf. Comput.
  Vision Pattern Recognit. (CVPR)}, 2015, pp. 3431--3440.

\bibitem{ronneberger2015u}
O.~Ronneberger, P.~Fischer, and T.~Brox, ``U-net: Convolutional networks for
  biomedical image segmentation,'' in \emph{Proc. Int. Conf. Med. Image Comput.
  Comput.-Assisted Intervention}.\hskip 1em plus 0.5em minus 0.4em\relax
  Springer, 2015, pp. 234--241.

\bibitem{chen2020simple}
T.~Chen, S.~Kornblith, M.~Norouzi, and G.~Hinton, ``A simple framework for
  contrastive learning of visual representations,'' in \emph{Int. Conf. Mach.
  Learn. (ICML)}.\hskip 1em plus 0.5em minus 0.4em\relax PMLR, 2020, pp.
  1597--1607.

\bibitem{oord2018representation}
A.~v.~d. Oord, Y.~Li, and O.~Vinyals, ``Representation learning with
  contrastive predictive coding,'' \emph{arXiv preprint arXiv:1807.03748},
  2018.

\bibitem{khaertdinov2021deep}
B.~Khaertdinov, E.~Ghaleb, and S.~Asteriadis, ``Deep triplet networks with
  attention for sensor-based human activity recognition,'' in \emph{Int. Conf.
  Pervasive Comput. Commun. (PerCom)}.\hskip 1em plus 0.5em minus 0.4em\relax
  IEEE, 2021, pp. 1--10.

\bibitem{wang2021exploring}
W.~Wang, T.~Zhou, F.~Yu, J.~Dai, E.~Konukoglu, and L.~Van~Gool, ``Exploring
  cross-image pixel contrast for semantic segmentation,'' in \emph{Proc. IEEE
  Int. Conf. Comput. Vision (ICCV)}, 2021, pp. 7303--7313.

\bibitem{khosla2020supervised}
P.~Khosla, P.~Teterwak, C.~Wang, and et~al., ``Supervised contrastive
  learning,'' \emph{Adv. neural inf. proces. syst. (NeurIPS)}, vol.~33, pp.
  18\,661--18\,673, 2020.

\bibitem{deldari2021time}
S.~Deldari, D.~V. Smith, H.~Xue, and F.~D. Salim, ``Time series change point
  detection with self-supervised contrastive predictive coding,'' in
  \emph{Proc. World Wide Web Conf. (WWW)}, 2021, pp. 3124--3135.

\bibitem{eldele2021time}
E.~Eldele, M.~Ragab, Z.~Chen, M.~Wu, C.~K. Kwoh, X.~Li, and C.~Guan,
  ``Time-series representation learning via temporal and contextual
  contrasting,'' \emph{Int. Joint Conf. Artif. Intell. (IJCAI)}, 2021.

\bibitem{farha2019ms}
Y.~A. Farha and J.~Gall, ``Ms-tcn: Multi-stage temporal convolutional network
  for action segmentation,'' in \emph{Proc. IEEE Comput. Soc. Conf. Comput.
  Vision Pattern Recognit. (CVPR)}, 2019, pp. 3575--3584.

\bibitem{van2018representation}
A.~Van~den Oord, Y.~Li, and O.~Vinyals, ``Representation learning with
  contrastive predictive coding,'' \emph{arXiv e-prints}, pp. arXiv--1807,
  2018.

\bibitem{kalantidis2020hard}
Y.~Kalantidis, M.~B. Sariyildiz, N.~Pion, P.~Weinzaepfel, and D.~Larlus, ``Hard
  negative mixing for contrastive learning,'' \emph{Adv. neural inf. proces.
  syst. (NeurIPS)}, vol.~33, pp. 21\,798--21\,809, 2020.

\bibitem{abedin2021attend}
A.~Abedin, M.~Ehsanpour, Q.~Shi, H.~Rezatofighi, and D.~C. Ranasinghe, ``Attend
  and discriminate: Beyond the state-of-the-art for human activity recognition
  using wearable sensors,'' \emph{Proc. ACM Interact. Mob. Wearable Ubiquitous
  Technol. (IMWUT)}, vol.~5, no.~1, pp. 1--22, 2021.

\bibitem{wei2016convolutional}
S.-E. Wei, V.~Ramakrishna, T.~Kanade, and Y.~Sheikh, ``Convolutional pose
  machines,'' in \emph{Proc. IEEE Conf. Comput. Vis. Pattern Recognit. (CVPR)},
  2016, pp. 4724--4732.

\bibitem{reiss2012introducing}
A.~Reiss and D.~Stricker, ``Introducing a new benchmarked dataset for activity
  monitoring,'' in \emph{Proc. Int. Symp. Wearable Comput. (ISWC)}.\hskip 1em
  plus 0.5em minus 0.4em\relax IEEE, 2012, pp. 108--109.

\bibitem{hanley1983method}
J.~A. Hanley and B.~J. McNeil, ``A method of comparing the areas under receiver
  operating characteristic curves derived from the same cases.''
  \emph{Radiology}, vol. 148, no.~3, pp. 839--843, 1983.

\end{thebibliography}

\end{document}